\documentclass[twocolumn, 10pt, times, A4paper]{article}
\usepackage{hhline}
\usepackage{booktabs}
\usepackage[table,xcdraw]{xcolor}
\usepackage{multirow}
\usepackage{geometry}
\usepackage{xcolor}

\usepackage{hyperref}
\hypersetup{colorlinks = true,
            linkcolor = red,
            urlcolor  = blue,
            citecolor = green,
            anchorcolor = blue}
\usepackage{mathptmx}
\usepackage[justification=centering]{caption}

\usepackage{threeparttable}
\usepackage{booktabs, caption, makecell}
\usepackage{enumitem}

\usepackage{natbib}
\usepackage{graphicx}
\usepackage[table,xcdraw]{xcolor}
\usepackage{amsmath, amssymb}
\usepackage{epsfig}
\usepackage{graphicx}
\usepackage{vmargin}
\usepackage{url}
\usepackage{tabularx}
\usepackage{boxedminipage}

\usepackage{newtxtext}

\usepackage{lastpage} 
\usepackage{fancyhdr}
\usepackage{soul} 

\newcommand{\captionfonts}{\fontsize{9}{9}\it}
\makeatletter 
\long\def\@makecaption#1#2{ %
 \vskip\abovecaptionskip
 \sbox\@tempboxa{{\captionfonts #1: #2}}%
 \ifdim \wd\@tempboxa >\hsize
	{\centering\captionfonts #1: #2\par}
 \else
 \hbox to\hsize{\hfil\box\@tempboxa\hfil}%
 \fi
 \vskip\belowcaptionskip}
\makeatother 
 
\vfuzz=\hfuzz

\newlength{\defbaselineskip}
\newcommand{\setlinespacing}[1]%
   {\setlength{\baselineskip}{#1 \defbaselineskip}}

\makeatletter
\newsavebox\saved@arstrutbox
\newcommand*{\setarstrut}[1]{%
	\noalign{%
		\begingroup
		\global\setbox\saved@arstrutbox\copy\@arstrutbox
		\global\setbox\@arstrutbox\hbox{%
			\vrule \@height #1
			\@depth 0cm
			\@width\z@
		}%
		\endgroup
	}%
}
\newcommand*{\restorearstrut}{%
	\noalign{%
		\global\setbox\@arstrutbox\copy\saved@arstrutbox
	}%
}
\makeatother
\newcommand{\paddingtop}[2]{\setarstrut{#1} #2 \\ \restorearstrut}

\makeatletter

\renewcommand\section{\@startsection{section}{1}{\z@}{6pt}{3pt}{\normalfont\large\bfseries}}
\renewcommand\subsection{\@startsection{subsection}{1}{\z@}{6pt}{3pt}{\normalfont\normalsize\bfseries}}

\renewcommand\subsubsection{\@startsection{subsubsection}{1}{\z@}{6pt}{3pt}{\normalfont\normalsize\itshape}}

\makeatother

\newcommand{\authorfont}{\fontsize{11}{14}\selectfont}
\newcommand{\titlefont}{\fontsize{12}{14}\selectfont \bf}

\newcommand{\tablefont}{\fontsize{9}{9}\selectfont}

\begin{document}

\bibliographystyle{apsr}
\date{}
\title{\vspace{-9mm} \titlefont NHA12D: A NEW PAVEMENT CRACK DATASET AND A COMPARISON STUDY OF CRACK DETECTION ALGORITHMS
 \vspace{-5.5mm}
}
\author{
\authorfont{~}\\ 
\authorfont{Zhening Huang$^1$, Weiwei Chen$^1$, Abir Al-Tabbaa$^1$, Ioannis Brilakis$^1$}\\
\authorfont{$^1$Department of Engineering, University of Cambridge}\\
\authorfont{~}\\ 
\vspace{-15mm}
}

\maketitle

\thispagestyle{empty}

\setul{2pt}{0.25pt} 

\section*{Abstract}
\addtocounter{section}{1}
Crack detection plays a key role in automated pavement inspection. Although a large number of algorithms have been developed in recent years to further boost performance, there are still remaining challenges in practice, due to the complexity of pavement images. To further accelerate the development and identify the remaining challenges, this paper conducts a comparison study to evaluate the performance of the state of the art crack detection algorithms quantitatively and objectively. A more comprehensive annotated pavement crack dataset (NHA12D) that contains images with different viewpoints and pavements types is proposed. In the comparison study, crack detection algorithms were trained equally on the largest public crack dataset collected and evaluated on the proposed dataset (NHA12D). Overall, the U-Net model with VGG-16 as backbone has the best all-around performance, but models generally fail to distinguish cracks from concrete joints, leading to a high false-positive rate. It also found that detecting cracks from concrete pavement images still has huge room for improvement. Dataset for concrete pavement images is also missing in the literature. Future directions in this area include filling the gap for concrete pavement images and using domain adaptation techniques to enhance the detection results on unseen datasets.

\section*{Introduction}
\begin{figure}[ht] 
	\centering
	\includegraphics[width=0.5\textwidth]{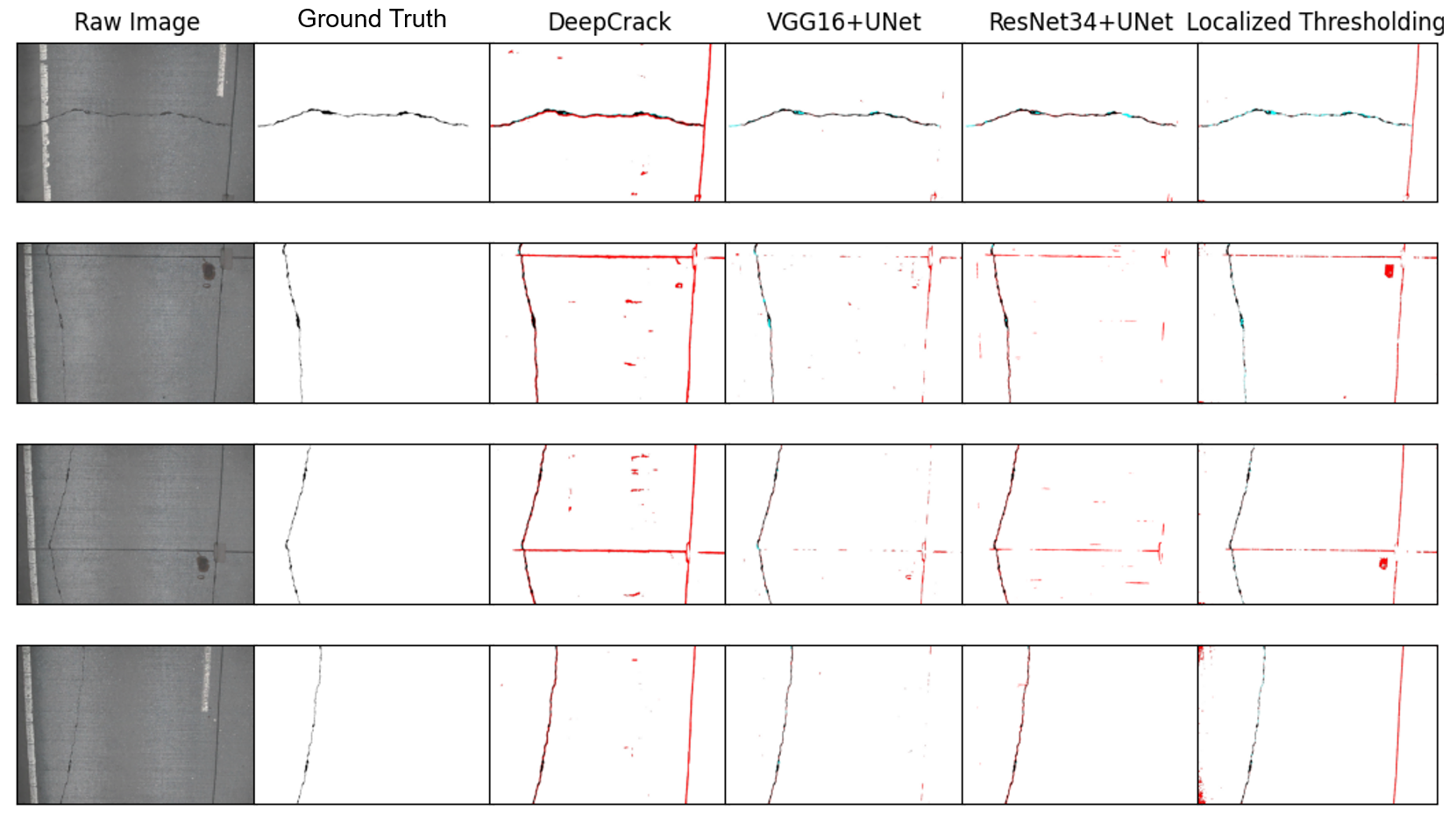}
	\vspace*{-6pt}
	\caption{Crack detection results on the proposed dataset. SoTA crack detection algorithms still have outstanding high false positive rates on concrete pavement images.\\ Black pixels: true positive; red pixels: false positive; green pixels: false negative}
	\label{fig:topimag}
\end{figure}

~
Automated pavement inspection involves taking RGB images of pavements with surveying vehicles and assessing their condition with image processing technologies. Crack detection is the task to extract crack information from 2D pavement images. There are various definitions of crack detection, and this paper focuses on pixel-level crack detection, which classifies each pixel in an image into the crack or non-crack pixel. Cracking is the most common type of surface distress on pavements and the severity of cracking on pavements is an essential indicator of pavement conditions (\cite{infrastructures3040058}). As a result, crack detection plays a key role in pavement management systems for determining optimal maintenance strategies. Crack detection from 2D images is a long-standing research challenge due to its inherent irregular patterns of crack, noise in images and different lighting conditions. In the early days, researchers used image processing techniques such as edge detection to extract the skeleton of cracks from images. In recent years, a large number of machine learning-based crack detection algorithms, such as DeepCrack ( \cite{Jenkins2018}), Crackforest(\cite{Shi2016}) have been developed and many of them demonstrated promising performance. However, to date, fully automatic crack detection remains a challenge as in practice, pavement images captured on highways have more complex backgrounds and conditions than images in public datasets. \\
To further accelerate the performance, we propose a new dataset, which contains images with more comprehensive cases, such as different pavement types and viewpoints. The pavement images in the proposed dataset were captured by National Highway’s digital survey vehicles on the A12 network, the UK. Cracks in the proposed dataset were visually identified and labelled at the pixel level. Moreover, a benchmark study for the state of the art (SOTA) crack detection algorithms is conducted, where three SOTA crack detection algorithms were selected and tested on the proposed dataset. The performance of three different detection models is quantitatively and objectively evaluated with consistent evaluation matrices. The detection results, as an example shown in Figure \ref{fig:topimag}, are used to identify the remaining challenges and potential directions for future development.\\
This paper is organized as follows. The section \textbf{Introduction} briefly introduces the background, research gap, and objectives for this study. The section “\textbf{Related Work}” describes the state of the art crack detection algorithms, existing benchmark studies and available pixel-level crack detection datasets. The section “\textbf{Methodology}” discusses the proposed dataset and implementation details for the comparison study. The section “\textbf{Results and Discussion}” evaluates the numerical and visual results of the models and discusses outcomes. The potential future research direction is also highlighted.
The section “\textbf{Conclusions}” presents the study conclusions and highlights recommendations for future research.

\section {Related Work}~
In the past two decades, methods to detect cracks from 2D pavement images have been investigated extensively as the need to achieve automated pavement inspection arises. In general, methods proposed by different scholars can be divided into two categories: Image processing-based methods and machine learning-based methods. Image processing-based methods use techniques such as intensity thresholding (\cite{Wang2011, Katakam2009, Oliveira2009, Li2016, Tang2013, Hu2010}), edge detection (\cite{Zhao2010, Attoh-Okine2008, Wang2007, Yan2007}) to extract crack skeleton from images. However, these methods tend to fail under complicated backgrounds and have low robustness against the change of environments. Machine learning-based methods, especially deep learning methods, although require a large amount of annotated datasets for training, have shown impressive results in recent years and outperform image processing-based methods in many dimension. 

\subsection{Crack Detection Algorithms}~

In the past five years (2016-2021), numerous machine learning algorithms were developed for crack detection. \cite{Shi2016} proposed a novel crack detection framework based on random structured forests named CrackForest. Some general algorithms for 2D image segmentation, such as UNet (\cite{Jenkins2018}), DeepLab (\cite{Chen2018}) have also been proven to be very powerful for crack segmentation. \cite{Lau2020} proposed a UNet based network architecture in which they replace the encoder with a pre-trained ResNet-34 neural network. Likewise, \cite{Dung2019} proposed a method based on a deep fully convolutional network (FCN) to detect cracks in concrete images. \cite{Jenkins2018} proposed DeepCrack CNN for automatic crack detection by learning high-level features for crack representation. \cite{Yang2020} proposed a novel network architecture, named Feature Pyramid and Hierarchical Boosting Network (FPHBN), for pavement crack detection. These methods have all been tested on public data and proved to have great results. However, most algorithms are only tested on one or two public datasets, while their robustness across multiple datasets remains unevaluated.

\subsection{Comparison Studies}
With the rapid emergence of new crack detection algorithms, the necessity to benchmark different crack detection algorithms also increases. In the literature, most model comparisons were conducted when a new algorithm was proposed (\cite{Yang2020, Zou2019, Lau2020, Li2019}). This type of comparison, however, tends to be inaccurate due to several reasons. For example, in some publications, different CNN-based models were evaluated without being trained on the same dataset, which makes the comparison unrigorous as the training scheme has sufficient impacts on the model performance. Also, the detection results in some papers were cited directly from other papers without considering the image pre-processing steps, implementation methods and evaluation details. Results from these comparisons tend to subject to bias to support the proposed algorithms. In the past, there are also some research work that specifically focused on model comparison. For example, \cite{Tsai2010} carried out a benchmark study for various crack detection models. In this work, six different algorithms were tested on pavement images captured on interstate highway I-75/I-85 near Atlanta and provided by the Georgia Department of Transportation. However, as this work was conducted in 2010, before the rise of machine learning-based methods, only image processing models were assessed. \cite{DORAFSHAN20181031} also tested six edge detection models and three deep learning models on concrete images. However, models used in that study detect cracks on classification level and cannot achieve pixel-level detection. \cite{Hsieh2020} carried out a benchmark for machine learning-based crack detection models, but their focuses were on 3D crack pavement images that were captured by laser scanners. Therefore, a comprehensive and robust comparison of the latest 2D crack detection algorithms is still missing in the literature.

\subsection{Crack Detection Dataset}
To fulfil the needs of benchmark study and training machine learning models, several crack detection datasets were proposed in the literature. A comprehensive search of the public crack dataset that labelled crack at the pixel level is conducted and the available datasets are listed in Table \ref{datasetsummary}. CFD dataset (\cite{Shi2016}) is the most widely used dataset for crack detection, which consists of 118 images of cracks on urban road surfaces in Beijing taken by iPhone 5. Each image is resized to $480 \times 320$ pixels and has been labelled at the pixel level. Similarly, Crack500 (\cite{Lau2020}) that were collected on the main campus of Temple University using cell phones, consists of 500 RGB images of pavement cracks of size around $2560 \times 1440$ pixels that. The other seven datasets in the literature, along with their parameters, are listed in table \ref{datasetsummary}. In summary, most datasets were captured on asphalt pavements, and images on concrete pavement is rare. Despite a large number of available datasets in the literature, the dataset that covers a range of complicated scenes on pavements is still needed. 


\begin{table*}[ht]
\centering
\caption{A comparison of existing crack dataset and the proposed NHA12D dataset}
\vspace{3pt}
\begin{threeparttable}[t]
\resizebox{\textwidth}{!}{%
\begin{tabular}{cccccc}
\hline \toprule[1pt]
\textsc{Name} &\textsc{Reference}  &\textsc{Resolution} &\textsc{Type of Label} &\textsc{Viewpoint} &\textsc{Surface} \\
\midrule[0.8pt]
CFD\tnote{1} & \cite{Shi2016} & $480\times 320$ & Pixel-Level & Vertical & Asphalt \\
Crack500 & \cite{Lau2020} & $2560 \times 1440$ & Pixel-Level & Random & Asphalt \\
CrackTree260 & \cite{Zou2012} & $960\times 720$ & Skeleton\tnote{2} & Vertical & Asphalt \\
DeepCrack & \cite{Zou2019} & $544\times 384$ & Pixel-Level & Random\tnote{3} & Random \\
GAPs384 & \cite{Yang2020} & $1920\times 1080$ & Patch-Level\tnote{4} & Vertical & Asphalt \\
Stone331 & \cite{Zou2019} & $1024\times 1024$ & Pixel-Level & Random & Stone \\
CRKWH100 & \cite{Zou2019} & $512\times 512$ & Pixel-Level & Vertical & Asphalt \\
CrackLS315 & \cite{Zou2019} & $512\times 512$ & Pixel-Level & Vertical & Asphalt \\
AigleRN & \cite{Chambon2011} & $991\times 462$ & Pixel-Level & Vertical & Asphalt \\
\textbf{NHA12D} & -- & $1920\times 1080$ & Pixel-Level & \textbf{Vertical+Forwarding} & \textbf{Concrete+Asphalt} \\ \bottomrule[2pt]

\end{tabular}%
}

\begin{tablenotes}
 \small
 \item[1] Most commonly used crack dataset to verify the detection results for crack detection models
 \item[2] CrackTree260 labels only depict the skeleton of the crack without rendering the width
 \item[3] DeepCrack contains a variety of surface including pavements, building walls, stone, metals and so on.
 \item[4] GAPs384 are divided into $64 \times 64$ patches and each patch is labelled as a crack or not
 \end{tablenotes}
\end{threeparttable}
\label{datasetsummary}
\end{table*}

\section{Methodologies}~
\subsection{Proposed Dataset}~
An annotated road crack dataset named NHA12D is proposed. The proposed dataset is directly taken by National Highways Surveying vehicles. This dataset is composed of 80 pavement images, including 40 concrete pavement images and 40 asphalt pavement images. Each image has a resolution of \(1920 \times1080\) and captures the view for an individual lane on the A12. Some images also contain vehicles, road markings, road studs, water stains, and other random objects. Concrete pavements have transversal and longitudinal joints every certain meter, which can be challenging for crack detection models to differentiate. For different types of pavements, 25 images were taken by cameras with a vertical viewpoint, and the other 15 were taken with a forwarding camera\footnote{Available at  \href{https://github.com/ZheningHuang/NHA12D-Crack-Detection-Dataset-and-Comparison-Study}{https://github.com/ZheningHuang/NHA12D-Crack-Detection-Dataset-and-Comparison-Study}}. 

\begin{figure*}[h] 
	\centering
	\includegraphics[width =\textwidth]{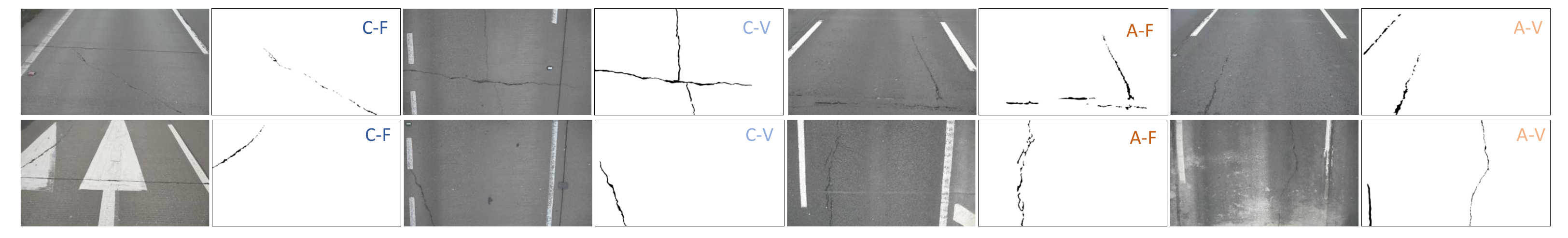}
	\caption{Our proposed Crack Detection Dataset (NHA12D) contains different pavement types and viewpoints images. Each image captures the view of entire lanes and provides the high-quality pixel-level crack annotation. \\C: Concrete, A: Asphalt, F: Forwarding view, V: Vertical view }
	\label{DHA12D}
\end{figure*}

Each image was carefully labelled at the pixel level. The segmentation masks are binary images where crack pixels have 255 as their intensity value while the other areas, i.e. backgrounds have intensity values of 0, as shown in Figure \ref{DHA12D}.

\subsection{Models selection}
Three crack detection models were selected for this benchmark study, including:
\begin{itemize}
    \item VGG-16/UNet (\cite{Dung2019}), a model that uses VGG-16 as the backbone encoder, UNet as the main architecture, and \textit{binary cross entropy} as loss function
    \item DeepCrack (\cite{Zou2019}), a CNN-based model that is designed specifically for crack segmentation 
    \item ResNet-34/UNet (\cite{Lau2020}), a model that has ResNet-34 as its backbone, UNet as main architecture, and the model is trained directly by optimizing the \textit{dice coefficient} 
\end{itemize}

\section{Implementation Details}

All training experiments in this project were performed on Intel Core i7-8700 Six-Core CPU and a single NVIDIA Tesla P100 16 GB GPU.  The implementation details for different models are listed below: 
\begin{itemize}
     \item\textbf{VGG-16/UNet:} The VGG-16/UNet architecture is built with implemented with Pytorch API. The weights of VGG-16 pre-trained on the ImageNet dataset is used for initialization.
     \item\textbf{DeepCrack:} The DeepCrack Model was also developed with Pytorch. The source code was published along with the paper and was used in this project for training and testing (\cite{Zou2019}). 
     \item\textbf{ResNet-34/UNet:} A model to implement ResNet-34/UNet was developed with Tensorflow/ Keras. The ResNet-34 encoder was pre-trained on ImageNet datasets.
\end{itemize}

\subsection{Training Scheme}
To optimize the performance of learning based models, we trained all models on a comprehensive dataset that contains all available crack datasets listed in table \ref{datasetsummary}. We also added some non-crack images for data balance purposes. Overall, this merged dataset contains 11,220 images, with the data distribution and pixel evaluation shown in Table \ref{tab:mergeddata}. All images were resized to \(448 \times 448\). 

\begin{table}[h]
\centering
\caption{Distribution of crack and Non-crack images in the merged dataset}
\vspace{3pt}
\label{tab:mergeddata}
\renewcommand{\arraystretch}{1}
\begin{tabularx}{1.0\columnwidth}{>{\centering}XXX}
  \hline
  \paddingtop{3pt}{&&}
  \textsc{Image Type} & \textsc{Number in Training Set} & \textsc{Number in Validation Set} \\
  \hline
  \paddingtop{3pt}{&&}
  \paddingtop{6pt}{&&}
   Crack Images  & \tablefont 8405 & \tablefont 1484  \\
  \paddingtop{3pt}{&&}
  \paddingtop{6pt}{&&}
  \tablefont{Non-Crack Images} & \tablefont 1199 & \tablefont 212 \\
  \hline
\end{tabularx}
\end{table}

\subsubsection{Data Augmentation}
Image augmentation artificially creates training images through different processing methods or a combination of multiple processing methods. It can enlarge the training dataset and therefore boost the performance of models. In this experiment, image augmentation was applied randomly to each image during the training process to increase the number of training examples virtually. Three types of augmentation were performed:
\begin{itemize}
    \item Rotate image randomly between 0$^{\circ}$ to 360 $^{\circ}$ 
    \item Random flips in horizontal and vertical axes
    \item Random change in lighting
\end{itemize}
The same augmentation was also applied to the ground truth of each training image, except for the random change in lighting. 

\subsection{Testing Scheme}
\subsubsection{Patchwise testing scheme}
Four crack detection model takes images input of size \(256 \times 256\). For large images that have a size of \(1920\times 1080\). This resizing process can significantly reduce the information stored in the images and therefore harm the testing results. To overcome this problem, we develop a new testing scheme that first crops the raw images into 24 small patches, which have sizes of \(270 \times 320\). Each patch was then tested independently with the detection models and the testing results were merged to form the final prediction. Another reason for cropping raw images is to compensate for the pixel ratio difference between testing and training datasets. As shown in table \ref{tab:pixelana}, for crack images in the training dataset, the crack pixel ratio is 4.66\%, while in the NHA12D dataset, this ratio is 0.40\% originally. Cropping images using the aforementioned method can significantly increase the ratio of crack pixels in crack images by around six times, to 2.40\%. Narrowing down the ratio difference between testing and training datasets has the potentials to boost the model performance. 

\begin{table}[h]
\centering
\caption{The percentage of crack and non-crack pixels in the NHA12D dataset, before and after the patch-wise pre-processing step}
\vspace{3pt}
\label{tab:pixelana}
\renewcommand{\arraystretch}{1}
\begin{tabularx}{1.0\columnwidth}{>{\centering}XXX}
  \hline
  \paddingtop{3pt}{&&}
  \textsc{Pre-processing } & \textsc{Crack Pixel} & \textsc{Non-crack Pixel } \\
  \paddingtop{6pt}{&&}
  \hline
  \paddingtop{3pt}{&&}
  \paddingtop{6pt}{&&}
   Before  & \tablefont 0.40\% & \tablefont 99.60\%  \\
  \paddingtop{3pt}{&&}
  \paddingtop{6pt}{&&}
  \tablefont{After} & \tablefont 2.40\% & \tablefont 96.60\% \\
  \paddingtop{3pt}{&&}
  \paddingtop{6pt}{&&}
  \hline
  \tablefont{Training Set} & \tablefont 4.66\% & \tablefont 95.34\% \\
  \hline
\end{tabularx}
\end{table}
\vspace*{-6pt}

\subsubsection{Otsu thresholding}
\begin{figure}[ht] 
	\includegraphics[width=0.5 \textwidth]{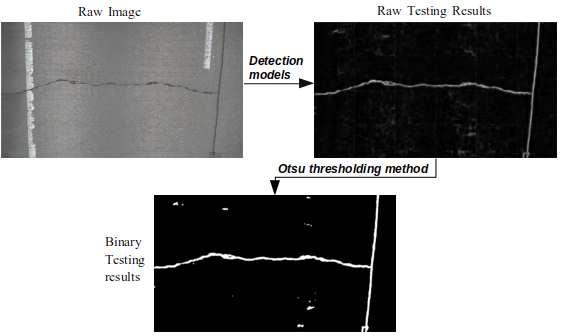}
	\caption{An illustration of using the Otsu threshold to binarize testing results.}
	\label{fig:binarytesting}
\end{figure}

The testing results from different models are greyscaled images that have pixel intensity varies from 0 to 255. The value of each pixel represents the likelihood of that pixel being a crack. However, the ground truth dataset is made up of binary images with the crack pixels being 255 and non-crack pixels being 0. To make them comparable, the Otsu thresholding method was applied to binarize the testing results. Otsu threshold returns a single intensity threshold that separates pixels into two classes, foreground and background, by minimizing intra-class intensity variance. An example of this operation is shown in Figure \ref{fig:binarytesting}. 

\begin{figure*}[ht!]
	\centering
  	\includegraphics[width=\textwidth]{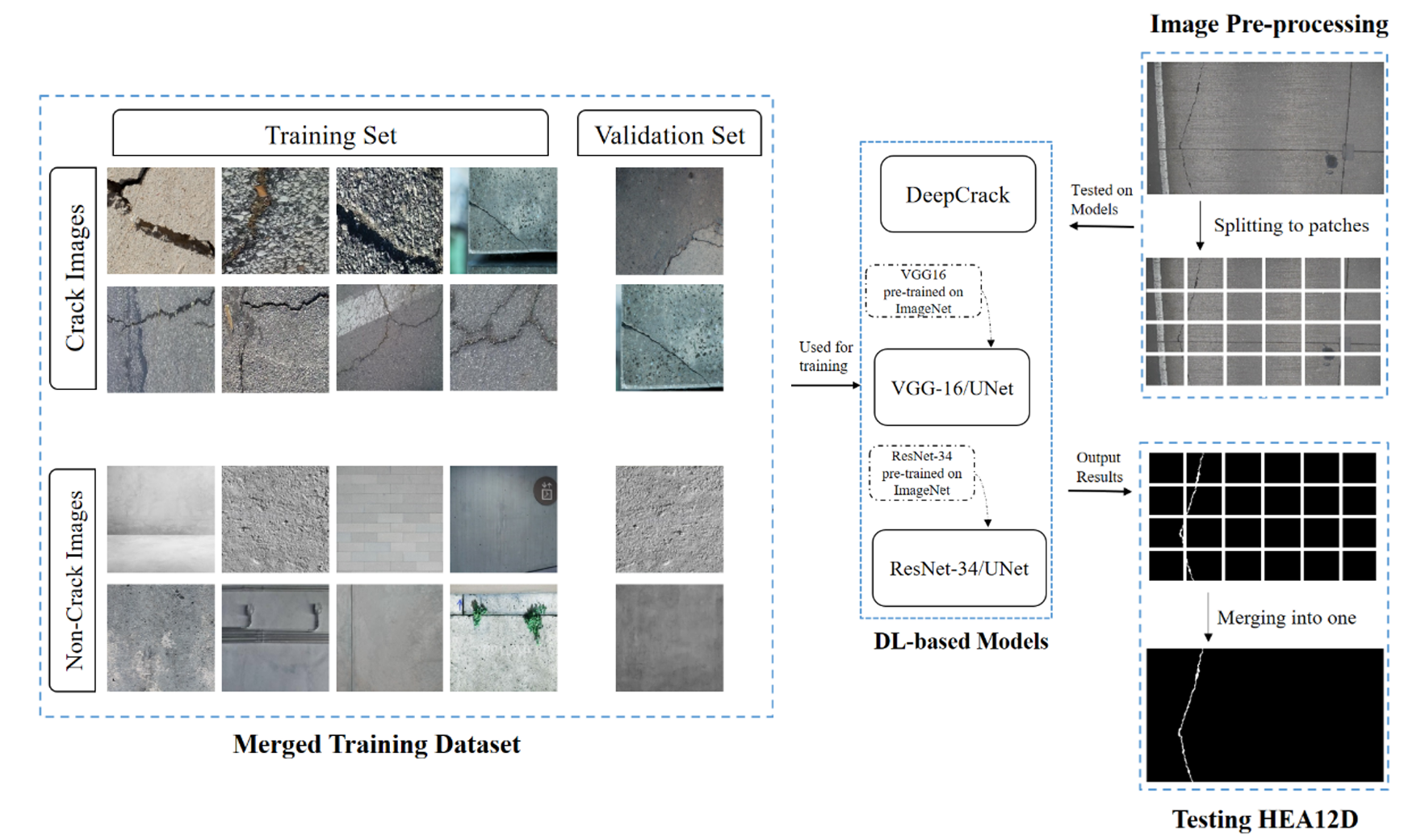}
  	\caption{The overall training and testing process of this benchmark study. The NHA12D were split into patches before being tested on three CNN-based models that were pre-trained on the merged dataset and the Localized Thresholding model. The tested patches were merged to form the final testing results.}
  	\label{fig:overall}
\end{figure*}

\subsubsection{Evaluation metrics}
Precision, Recall and F1 score are the three most commonly used parameters for crack detection evaluation, so we choose this metrics for this study. Three parameters can be obtained by using the following equations.  
\begin{equation*}
Precision= \frac{TP}{TP+FP} \qquad 
Recall= \frac{TP}{TP+FN}
\end{equation*}
\begin{equation*}
F_{1}=\frac{2\times TP}{2 \times TP+FP+FN} 
\end{equation*}
In the evaluation step, the value of TN, FP, TN, TP were calculated by comparing the binary testing results of images to their ground truth segmentation masks. Also because there are transition regions between the crack pixels and the non-crack pixels in the subjectively labelled ground truth, results that are two pixels near any labelled crack pixel are considered as true positives. This evaluation convention is widely used in other crack detection research ( \cite{Lau2020,Amhaz2016,Fan,Liu2019,Ai2018}).
The overall training and testing process are illustrated in Figure \ref{fig:overall}.

\section{Results and Discussion}
\textbf{Numerical Results}\\
The testing results of three models on different images were evaluated from various perspectives. All results are summarized in Table \ref{summaryresults}. From the result, the VGG-16/UNet model has the best overall performance, as the F1 score indicates. The second-best performance model is the ResNet34/UNet model. From The recall perspective, the DeepCrack model outperformed other models with a very high margin, while from the precision perspective, two UNet-based models demonstrated similar performance and outperform better than the other two models.
Models tested on concrete pavement images generally had higher recall scores, but lower precision scores, due to the false positive results on concrete joint region. 
\begin{figure*}[h!]
	\centering
  	\includegraphics[width=0.8\textwidth]{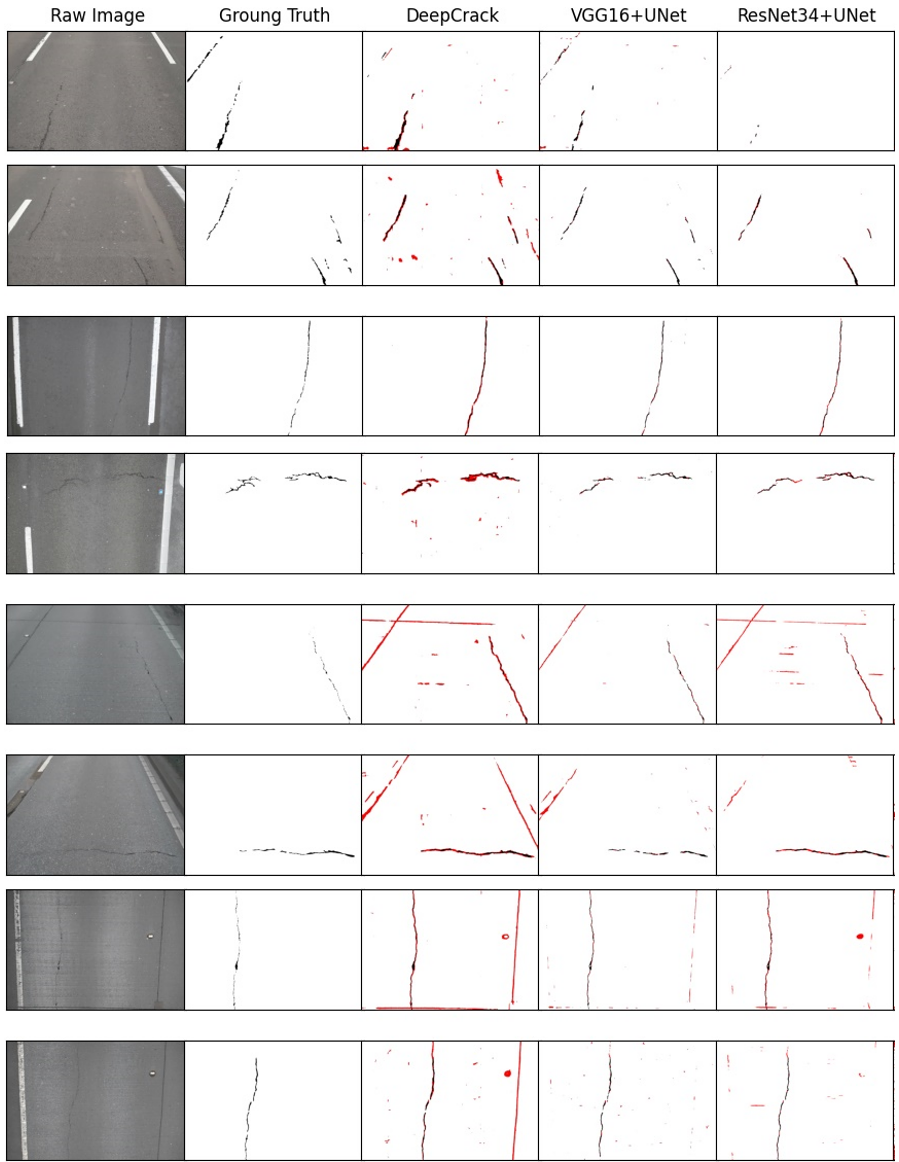}
  	\vspace{3pt}
  	\caption{Eight examples of crack detection results by different models, including 2 images for each pavement type on each viewpoint in NHA12D dataset.  \\Black pixels:
True Positive; Red pixels: False Positive, Green Pixels: False Negative}
  	\label{fig:visualreult}
\end{figure*}

\begin{table*}[h!]
\centering
\caption{Quantitative comparisons of Four Algorithms on the NHA12D dataset, with the highest value highlighted in bold}
\vspace{6pt}
\resizebox{1\textwidth}{!}{
\begin{tabular}{cccccccc}
\hline \toprule[1pt]

\multirow{2}{*}{\textsc{Pavement Type}} & \multirow{2}{*}{\textsc{Technique}} &\multicolumn{3}{c}{\textsc{Vertical View}} &\multicolumn{3}{c}{\textsc{Forwarding View}}\\ &
& {\cellcolor[HTML]{EFEFEF}\textsc{Recall}} & {\cellcolor[HTML]{EFEFEF}\textsc{Precision}} & {\cellcolor[HTML]{EFEFEF}\textsc{F1-Score}} & {\cellcolor[HTML]{EFEFEF}\textsc{Recall}} & {\cellcolor[HTML]{EFEFEF}\textsc{Precision}} & {\cellcolor[HTML]{EFEFEF}\textsc{F1-Score}}\\ \midrule[0.8pt]

\parbox[t]{2mm}{\multirow{3}{*}{\rotatebox[origin=c]{90}{Concrete}}} 
& \textit{DeepCrack} & \textbf{0.963} &	0.219&	0.349 & \textbf{0.966} &	0.141 &	0.243
 \\
& \textit{VGG16/UNet} & 0.915&	\textbf{0.507}& \textbf{0.638} & 0.924& \textbf{0.353}& \textbf{0.504}
 \\
& \textit{ResNet34/UNet} & 0.935 & 	0.427 &	0.570 & 0.922&0.250&0.385
 \\

 \midrule[0.8pt]
 \parbox[t]{2mm}{\multirow{3}{*}{\rotatebox[origin=c]{90}{Asphalt}}} 
& \textit{DeepCrack} & \textbf{0.903} &	0.356&	0.505 & \textbf{0.803} &	0.320 &	0.435
 \\
& \textit{VGG16/UNet} & 0.780&	\textbf{0.565}& \textbf{0.646} & 0.616&\textbf{0.557}&\textbf{0.570}
 \\
& \textit{ResNet34/UNet} & 0.733 & 	0.533 &	0.603 & 0.498&0.564&0.477
 \\
\toprule[1pt]
\end{tabular}
}
\label{summaryresults}
\end{table*}

\subsection{Visual Results}

Figure \ref{fig:visualreult} shows sample visual results from different models. In total, 8 images are presented, including two images for each pavement type on each viewpoint. Black and Red colours are used to indicate true positive pixels and false-positive pixels respectively.\\
The DeepCrack model generally captured all crack pixels in images regardless of viewpoints and pavement types, although it also commonly misrecognized concrete joints as cracks in all images. Visually, the VGG-16/UNet model has the closest match to the ground truth segmentation mask. The ResNet-34/UNet model demonstrated a very similar performance to the VGG-16/UNet model. The Localized thresholding model has reasonable performance on some images, but overall it failed to do the crack detection job on the NHA12D dataset.
\subsection{Discussion}
Considering that the purpose of using crack detection algorithms is to eliminate or sufficiently reduce the need to examine the raw images, it is important to capture most cracks and avoid misclassifying healthy surfaces. From this perspective, the three latest crack detection models perform poorly on the proposed pavement dataset. All models were unable to differentiate between cracks and concrete joints, leading to a high false-positive rate. Also, some model failed to recognize obvious cracks on asphalt pavements, making them unusable in real practice. The training dataset, which is a large collection of all available crack datasets, contains very few images for concrete pavements where joints and cracking co-exist. Consequently, such knowledge cannot transfer to model. This highlights the need to collect and label concrete pavement images for crack detection.  \\
Another reason for the unsatisfactory performance is the difference between the training datasets and testing datasets. The machine learning model normally assumes that the training and testing sets come from the same distributions/domain. Directly applying the trained model on the new dataset may cause degradation in the performance. In our case, models were trained on public crack datasets but then tested on new images in NHA12D dataset, which is a typical Domain Shift challenge. To overcome these challenges, using domain adaptation techniques such as generating synthetic data can boost performance. Research in this area is less studied in current literature. 
\section*{Conclusions}
Crack detection from 2D images has been investigated intensively and a large number of papers have been published in the last two decades. However, fully automated crack detection remains challenging due to the complex pavement images captured in practice. To thoroughly examine the SOTA crack detection algorithms, we proposed a new dataset that contains a more complicated background and various types of pavement. Also, we conducted a robust benchmark study of three the state of the art crack detection algorithms, including DeepCrack, VGG-16/UNet and Resnet34/UNet, on the proposed dataset, to quantitatively and objectively evaluate their performance. Experiments show that there is still a gap in concrete pavement crack detection because current models fail to separate concrete joints and cracks. From the outcome of the benchmark study, we identify the potential directions in the future including, a) producing the more well-labelled crack images on concrete pavements and developing algorithms for concrete pavement images, b) applying domain adaptation techniques to enhance the model performance on unseen images.
\newpage
\section*{References} 
\vspace{-4mm}
\bibliography{references}

\end{document}